\pdfoutput=1

\documentclass[11pt]{article}

\usepackage[final]{acl}

\usepackage{times}
\usepackage{latexsym}

\usepackage[T1]{fontenc}

\usepackage[utf8]{inputenc}

\usepackage{microtype}

\usepackage{inconsolata}

\usepackage{graphicx}

\usepackage{amsmath} 
\usepackage[ruled,linesnumbered]{algorithm2e}
\usepackage{fontawesome}
\usepackage{booktabs}
\usepackage{adjustbox}
\usepackage{cleveref}
\usepackage{array} 
\usepackage{booktabs} 

\title{When Emotional Stimuli meet Prompt Designing: \\An Auto-Prompt Graphical Paradigm}

\author{
 \textbf{Chengqian Ma\textsuperscript{1}},
 \textbf{Xiangyu Zhao\textsuperscript{1}},
 \textbf{Chunhui Zhang\textsuperscript{2}},
 \\
 \textbf{Yanzhao Qin\textsuperscript{3}},
 \textbf{Wentao Zhang\textsuperscript{3}},
\\
 \textsuperscript{1}Xiamen University,
 \textsuperscript{2}Dartmouth College,
 \textsuperscript{3}Peking University,
\\
}

\begin{document}
\maketitle

\begin{abstract}
With the development of Large Language Models (LLM), numerous prompts have been proposed, each with a rich set of features and their own merits. This paper summarizes the prompt words for large language models (LLMs), categorizing them into stimulating and framework types, and proposes an Auto-Prompt Graphical Paradigm(APGP) that combines both stimulating and framework prompts to enhance the problem-solving capabilities of LLMs across multiple domains, then exemplifies it with a framework that adheres to this paradigm. The framework involves automated prompt generation and consideration of emotion-stimulus factors, guiding LLMs in problem abstraction, diversified solutions generation, comprehensive optimization, and self-verification after providing answers, ensuring solution accuracy. Compared to traditional stimuli and framework prompts, this framework integrates the advantages of both by adopting automated approaches inspired by APE work, overcoming the limitations of manually designed prompts. Test results on the ruozhiba and BBH datasets demonstrate that this framework can effectively improve the efficiency and accuracy of LLMs in problem-solving, paving the way for new applications of LLMs.


\begin{figure}[ht]
    \includegraphics[width=\linewidth]{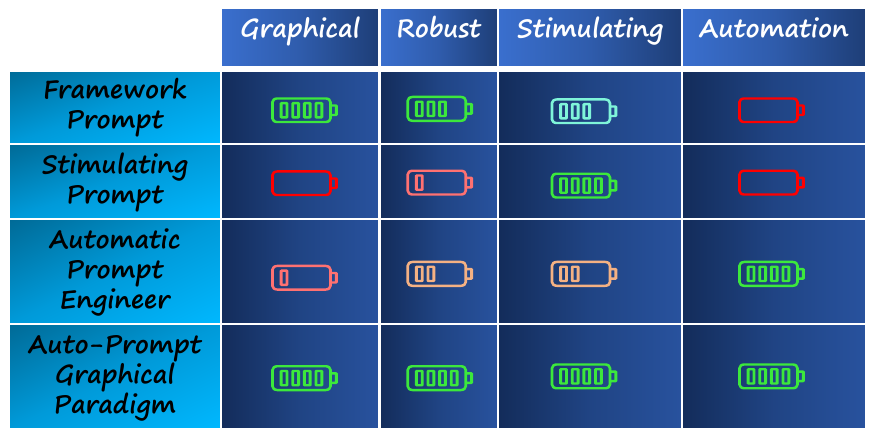}
	\caption{Performance comparison between traditional prompt and the Auto-Prompt Graphical Paradigm}
        \label{fig:fire}
\end{figure}

\end{abstract}  
\section{Introduction}
\label{sec:intro}
\begin{figure*}[t]
    \includegraphics[width=\linewidth]{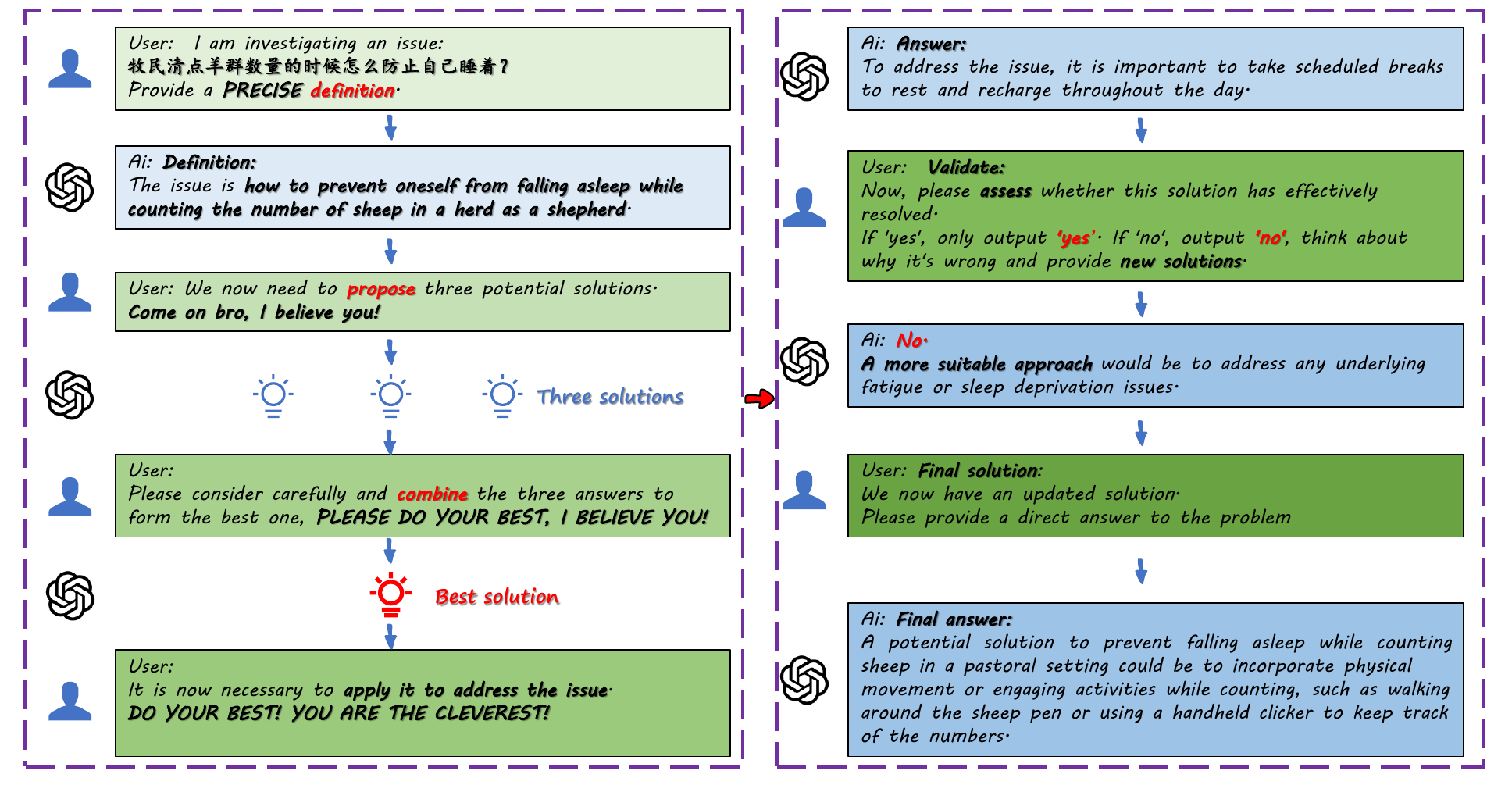}
	\caption{An example for using APGP to solve a problem, which in Chinese is "How to prevent falling asleep when counting the number of sheep for the herder? ". The gradually deepening blue boxes in the picture are the answers from the AI, and the green boxes are the guiding prompt words from the auto-prompt graphical paradigm. Each pair of blue and green boxes represents an interaction with the AI, with the execution order being to execute the one on the left first and then the one on the right.}
        \label{fig:example}
\end{figure*}

Since the advent of large language models, they have helped humanity solve numerous problems, liberating many from mundane tasks. Consequently, efforts have been made to leverage these models to tackle challenges that are difficult for humans, yielding a series of achievements. Large language models not only demonstrate an understanding of human language but also, by virtue of this understanding, offer insights into the world's knowledge underlying language. As a result, they have been applied to address problems beyond text and across multiple modalities.

In the quest to unleash the potential of large language models, CoT~\cite{wei2022nips} introduced the concept of progressive reasoning, which advocates for the gradual engagement of these models in cognitive processes. This idea has been inherited and evolved by subsequent works such as PS-CoT~\cite{wang-etal-2023-plan}, ToT~\cite{yao2023tree}, and GoT~\cite{Besta2023GraphOT}, expanding the cognitive architecture of large language models and rendering it highly flexible. Specifically, PS-CoT~\cite{wang-etal-2023-plan} extends a single CoT into multiple paths, while the backtracking mechanism proposed by ToT~\cite{yao2023tree} endows large language models with fault tolerance during problem-solving. Additionally, GoT~\cite{Besta2023GraphOT} introduces a diverse range of selectable operations for large language models in the problem-solving process. These problem-solving frameworks are all implemented through prompts, categorized as \textbf{Framework Prompt}. However, these approaches necessitate manual prompt design for each operation. To address the challenge of prompt design, the APE~\cite{zhou2023large} work proposes an automated prompt design method which entrusts the design of prompts to the LLMs. Unfortunately, this method fails to guide large language models to utilize flexible structures like those in GoT.

Beyond the problem-solving frameworks of large language models, numerous studies have indicated that these models exhibit some human-like characteristics. "Let’s think step by step" encourages large language models to consider problems more meticulously~\cite{Kojima2022LargeLM}, whereas "Take a deep breath and work on this problem step-by-step" can enhance the performance of large language models even better~\cite{yang2024large}. Large language models demonstrate positive responses to prompts encouraging encouragement, emphasis, threats, and other types~\cite{Li2023LargeLM}. This suggests that large language models trained on human corpora exhibit better responses to instructions containing human emotions, further highlighting their sensitivity as multi-modal tools for language and underlying world knowledge. Moreover, emotional stimuli play significant roles in decision-making, competitive sports performance~\cite{Lazarus2000How}, academic domains~\cite{Pekrun2002AcademicEI}, and other areas, broadening the application of large language models. These essential prompts are implemented through prompts, categorized as \textbf{Stimulating Prompt}.

The problem-solving framework prompts guide large language models, yet  their generalizability is constrained by task-specific characteristics. Stimulating prompts that leverage the human-like emotional characteristics of large language models often do not need to be altered based on tasks. However, they cannot provide sufficiently comprehensive frameworks for large models to solve problems effectively. The combination of the universality of stimulating prompts and the task-specific features of framework prompts can more effectively exploit the latent capabilities of large language models.

Combining the characteristics of Stimulating prompts and Framework prompts, we integrate the two while addressing the limitations of framework prompts. Referring to the APE approach~\cite{zhou2023large}, we propose a universal \textbf{auto-prompt graphical paradigm(APGP)} that considers human emotional stimuli and incorporates an automatic prompt-filling function which can automatically fill in the prompts required by the Framework Prompt. This paradigm aims to enhance the ability of large language models to solve problems across multiple domains. Subsequently, we provided a auto-prompt graphical framework to prove this paradigm.

Our main contributions are as follows:

\begin{itemize}
\setlength{\itemsep}{0pt}
\setlength{\parsep}{0pt}
\setlength{\parskip}{0pt}
    \item We integrated traditional prompts into two categories: Stimulating Prompt and Framework Prompt, then devised a new type of framework prompts with emotional stimuli, combining the advantages of both traditional prompt types.
    \item We designed an auto-prompt graphical paradigm(APGP) using the new type of prompt, which signifies a brand-new paradigm for prompt utilization. Then we offered a framework to confirm this paradigm.
    \item We tested the framework on datasets such as ruozhiba~\cite{Bai2024COIGCQIAQI} and BBH~\cite{Suzgun2022ChallengingBT}, yielding favorable results. Furthermore, we conducted ablation experiments to demonstrate the effectiveness of our approach.
\end{itemize}

\section{Related Work}
\label{sec:Related Work}
\subsection{Prompt-based LLM Reasoning}
In the realm of advancing large language models (LLMs), pioneering frameworks like Chain-of-Thought Prompting (CoT), Plan-and-Solve Prompting (PS), "Tree of Thoughts" (ToT), and Graph of Thoughts (GoT)~\cite{wei2022nips,wang-etal-2023-plan,yao2023tree,Besta2023GraphOT} have revolutionized problem-solving capabilities. Chain-of-Thought Prompting (CoT)~\cite{wei2022nips} enhances the problem-solving abilities of large language models (LLMs) by guiding them to simulate human-like step-by-step reasoning processes in their input prompts, resulting in more accurate and coherent outputs even without task-specific training. Plan-and-Solve Prompting (PS) ~\cite{wang-etal-2023-plan} improves large language models' (LLMs) performance on multi-step reasoning tasks, outperforming Zero-shot-CoT and rivaling manually-guided CoT methods, highlighting LLMs' potential for reasoning without manual examples. The "Tree of Thoughts" (ToT)~\cite{yao2023tree} framework empowers large language models to make thoughtful decisions by exploring multiple reasoning paths and self-assessing decisions within a tree-like structure of coherent textual units. Graph of Thoughts (GoT) ~\cite{Besta2023GraphOT} models the reasoning process of large language models (LLMs) as an arbitrary graph structure, significantly improving LLMs' performance on complex tasks. It enhances task quality and reduces costs, demonstrating advantages in various real-world applications and advancing LLMs' reasoning capabilities towards human-like thinking patterns.

\subsection{Emotion-Enhanced LLM Reasoning}
In recent studies, researchers have explored various techniques to improve the capabilities of large language models (LLMs). For instance, \cite{Li2023LargeLM} examines how large language models (LLMs) understand and respond to emotional stimuli, demonstrating their capability to comprehend emotional intelligence and improve performance with emotion prompts, paving the way for enhanced interaction between LLMs and human emotional intelligence. "Step-Back Prompting"~\cite{Zheng2023TakeAS} enhances large language models' (LLMs) reasoning capabilities in complex tasks by guiding them through abstract thinking processes, resulting in notable performance improvements across diverse challenging tasks such as STEM, knowledge question answering, and multi-hop reasoning. Chain-of-Verification (CoVe) method aims to mitigate the occurrence of "hallucination" in generated text by large language models, where seemingly plausible but factually incorrect information is produced. The "Rephrase and Respond" (RaR) ~\cite{Deng2023RephraseAR} method aims to enhance large language models' (LLMs) comprehension and responses to questions by allowing them to autonomously rephrase and expand posed questions, resulting in improved accuracy. 
\subsection{Graph-Dependency LLM Reasoning}
The Automatic Prompt Engineer (APE)~\cite{zhou2023large} method enhances large language models (LLMs) by automatically generating and selecting prompts. It outperforms previous LLM baselines on various tasks and approaches human-generated prompts' performance. Extensive experiments demonstrate that prompts generated by APE outperform previous LLM baselines on 24 out of 24 instruction induction tasks and 17 out of 21 curated BIG-Bench tasks, reaching or approaching the performance of prompts generated by human annotators.

\section{Methodology}
\label{sec:Methodology}
\subsection{Overview}
\begin{figure}[ht]
    \includegraphics[width=\linewidth]{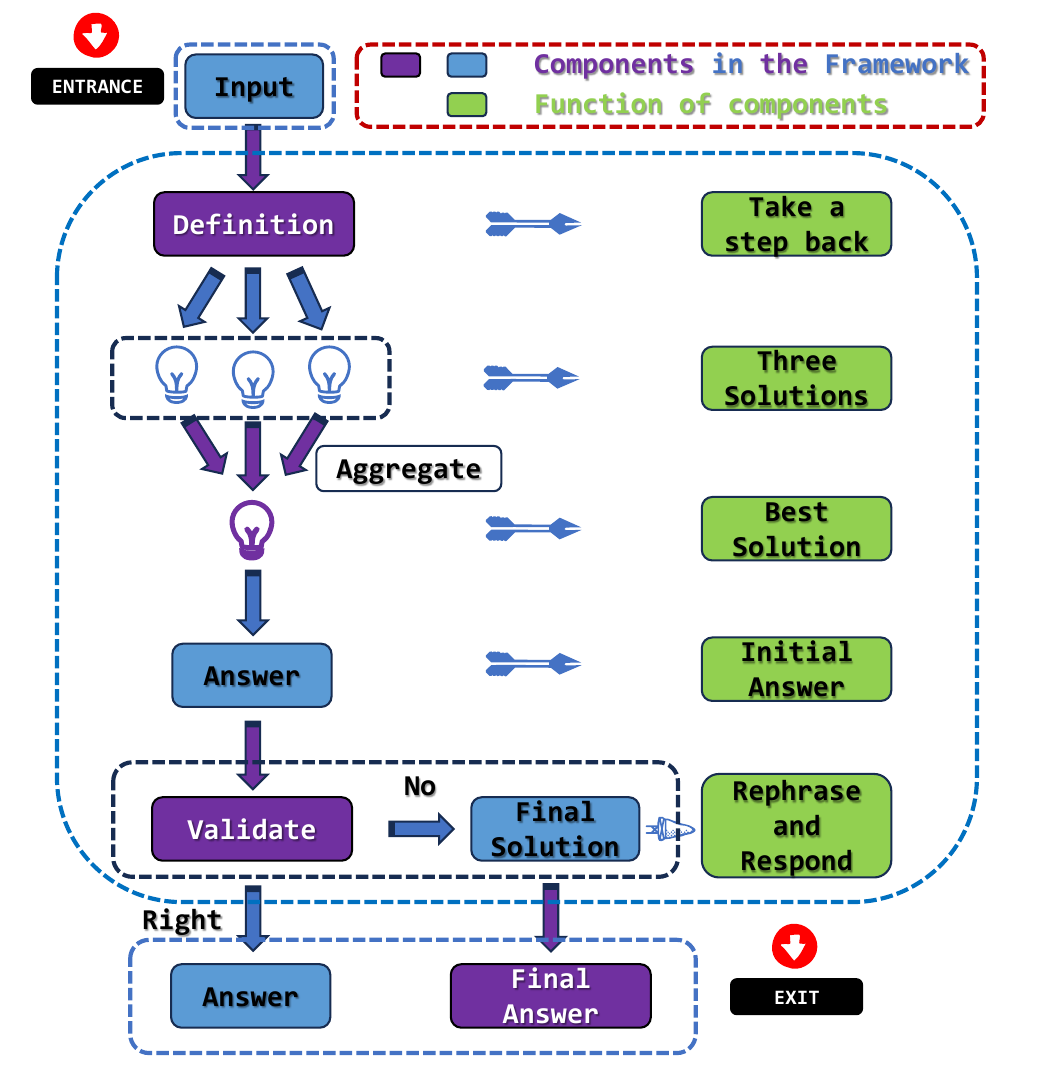}
	\caption{A Schematic Representation of the Stimuli Graphical Process within a Problem-Solving Framework.}
        \label{fig:flow_chart}
\end{figure}

As illustrated in \cref{fig:flow_chart}, our framework is a prompt-free approach, which not implying the absence of prompts, but rather eliminating the need for manually designed prompts to address problems.

The prompts in the framework consist of two parts: The first part guides the LLM in analyzing the problem, establishing the framework's structure, directing the LLM to propose different research approaches for different problems, and assessing the LLM's responses to determine the next course of action. The second part of the prompts can be provided by the LLM under sufficient emotional encouragement and guidance.

We will refer to the first part of the prompt as the immutable component, and refer to the prompt given by the LLM in the second part as the variable component. In the construction of the immutable components, termed as "fixed prompts", careful consideration was given to the diverse responses of the LLMs to different emotional stimuli. Notably, LLMs exhibit significantly positive responses to prompts related to encouragement and praise. Consequently, in the fixed prompts, a friendly and encouraging tone was adopted wherever feasible. Moreover, considering the LLMs' inclination towards exclamation marks and capitalized words, special attention was paid to the design of prompts requiring emphasis, employing relevant techniques to enhance their effectiveness.

\begin{algorithm}[ht]
\small
\caption{Stimuli Graphical Processor}\label{alg: Prccessor}
\KwIn{Description of the problem $\mathbf{P}_{desc}$}
\KwOut{The answer to the question $\mathbf{Ans}$}

\tcp{Abstract the problem}
$\mathbf{P}_{def} \gets \text{Definite}(\mathbf{P}_{desc})$

\tcp{Generate three solutions}
$\mathbf{S}_1, \mathbf{S}_2, \mathbf{S}_3 \gets \text{Generator}(\mathbf{P}_{def})$ 

\tcp{Aggregrate the best solution}
$\mathbf{S}_{Best} \gets \text{Aggregator}(\mathbf{P}_{def}, \mathbf{S}_1, \mathbf{S}_2, \mathbf{S}_3)$ 

\tcp{Get answer by the solution}
$\mathbf{Ans} \gets \text{Get\_Answer}(\mathbf{P}_{def}, \mathbf{S}_{Best})$

\tcp{Validate the answer}
$ Success\_Flag, \mathbf{S}_{Final} \gets \text{Validate}(\mathbf{P}_{def}, \mathbf{Ans})$

\eIf{Success\_Flag}{
\Return $\mathbf{Ans}$
      }{
      $\mathbf{Ans}_{Final} \gets \text{Get\_Answer}(\mathbf{P}_{def}, \mathbf{S}_{Final})$
      \Return $\mathbf{Ans}_{Final}$
      }

\end{algorithm}

As shown in \cref{alg: Prccessor}, our method consists of the following steps:

\begin{enumerate}
\item After receiving the problem, the first step is to prompt the LLM to provide a clear definition of the problem. This method, inspired by the "Take a Step Back" and "Rephrase and Respond" approaches, allows the large model to have a clearer understanding of the problem. By analyzing the problem at an abstract level and providing advanced guidance, the LLM can then proceed to solve the problem.
\item Once the formal definition of the problem is obtained, the LLM is tasked with generating three potential solutions. Here, we employ the conventional generation operation of chain-of-thought, generating multiple solutions to ensure the fault tolerance of the LLM. If one of the generated multiple solutions is obviously poor, or the shortcomings of one solution can be made up for by other solutions, then they can learn from each other.
\item After obtaining three potential solutions, the LLM combines them to generate the best solution, leveraging the strengths of each solution. Unlike traditional framework prompt methods that use scoring or voting to select the best solution, we recognize that thoughts can often reference and complement each other, collectively forming the optimal solution. This introduces a new thought aggregation approach.
\item With the final solution obtained, the large model then utilizes it to address the problem and provide an answer.
\item After obtaining the answer, the model is required to validate it, carefully considering whether the answer is correct. Generally speaking, as the number of parameters of LLM increases, its performance will become better and better, but for some lesser known torso and tail distribution~\cite{Sun2023HeadtoTailHK} facts, LLM does not have a correct understanding of them, but will construct some text that appears reasonable but is actually wrong, which is the hallucination phenomenon of LLM. This validation process, inspired by the CoVe approach in stimulus-based prompting, reduces the likelihood of hallucinations by the LLM.
\item Upon observing the validation results, if the LLM successfully solves the problem, the answer is outputted. If the validation results indicate failure, the LLM is prompted to generate new solutions based on the erroneous solution and then readdress the problem accordingly. This approach borrows from the "TP"~\cite{Yu2023ThoughtPA} technique, leveraging past experiences to aid in resolving current issues. If the validation process itself fails, the answer obtained in step four is returned directly.
\end{enumerate}

\subsection{Definition}
When tackling complex problems with LLMs, multiple steps are often required~\cite{wei2022nips}. However, errors occurring at any stage of this process can easily accumulate throughout the multiple steps~\cite{Yu2023ThoughtPA}, potentially leading to catastrophic consequences. To mitigate this, we employ the "take a step back"~\cite{Zheng2023TakeAS} technique, which involves first rephrasing the problem and simplifying it through a basic abstraction. This abstraction allows the LLM to grasp the essence of the problem, disregarding its details to avoid potential issues caused by intricacies.

\subsection{Get solutions and Aggregate}
In works like Plan-and-Solve~\cite{wang-etal-2023-plan}, CoT~\cite{wei2022nips} at.al., we observed that prompting LLMs to propose problem-solving plans before implementing them can stimulate their problem-solving abilities.

When prompting LLMs to devise solutions, we require them to propose three different approaches. This multiplicity of options provides LLMs with more choices when solving problems.

After obtaining three distinct proposals, our framework mandates LLMs to merge these approaches. Traditional aggregate is divided into two types: vote and score, among which vote is for the LLM to vote and select the thought with the most votes; while score is to score multiple thoughts, and select the thought with the highest score after sorting the scores. The vote method used by the traditional aggregate method may encounter the situation of a tie, and dealing with the situation of a tie requires extra tokens; the score method used by the traditional aggregate requires the LLM to be more sensitive to numbers, and the processing of numbers has always been the weakness of the LLM. Unlike conventional methods such as GoT~\cite{Besta2023GraphOT}, which directly filter results through voting or scoring operations—limited by LLMs' numerical abilities—our framework's merging operation entails LLMs synthesizing the advantages and disadvantages of the three proposals to form a comprehensive solution. This approach resembles biological hybridization~\cite{Dobzhansky1937GeneticsAT}, leveraging the diversity and distinct focuses of different methods. It's a brand-new type of aggregation operation.

\subsection{Get solutions and Validate}
After obtaining the answer, our framework does not rush to output it. Instead, we utilize the LLM's capability to evaluate the answer, verifying if it successfully resolves the problem. In social sciences, due to differing subjective experiences, the perception of the same event may be biased. Similar to ensuring mutual understanding through repetition in human conversation, we adopt this idea by having the LLM validate the previously obtained answer. This effectively creates two independent LLMs, akin to aligning perceptions between speakers and listeners. A positive validation indicates that the answer transcends individual perspectives, garnering broader acceptance.

Upon successful validation, we directly output the answer. Otherwise, we extract the experience from the incorrect answer and generate a better solution, using it to derive a new response. Here, we draw inspiration from TP methodology, employing past experiences in LLM tasks to aid in current problem-solving. In abstract terms, this also realizes the functionality of backtracking during traversal processes in graph structures: On the graph structure, it returns to the node of getting the answer.

\section{Experiments}
\label{sec:Experiments}
Through this framework, we applied the GPT-3.5-turbo model to the Ruozhiba~\cite{Bai2024COIGCQIAQI} and BIG-Bench Hard~\cite{Suzgun2022ChallengingBT} datasets. The framework's usage involves iterative interactions with the LLM, where each interaction consists of feeding text input to the LLM and parsing the LLM's response.
\subsection{Datasets}
\paragraph{Ruozhiba.}
The Ruozhiba dataset is a distinctive Chinese natural language processing dataset originating from the "Ruozhiba" community on Baidu Tieba, a Chinese online forum where members exchange ideas that are both peculiar and tinged with logic, filled with a wealth of brain teasers and metaphorical phrases. It consists of 500 post titles with the most likes, from which instructional prompts are selected, filtering out declarative or unanswerable content as well as harmful information. For these prompts, replies generated by humans or GPT-4 are collected, and GPT-4's responses are manually reviewed to ensure accuracy, resulting in 240 sets of high-quality (question, response) pairs. These data contain elements such as puns, polysemy, causal inversion, and homophones, designed with logical traps that pose challenges to both humans and AI. Due to its uniqueness and complexity, the Ruozhiba dataset demonstrates tremendous potential in enhancing AI models' logical reasoning and understanding of complex Chinese language structures. Experiments have shown that LLMs fine-tuned on the Ruozhiba dataset exhibit exceptionally superior performance.
\paragraph{BIG-Bench-Hard.} 
The BIG-Bench Hard (BBH) subset is derived from the original BIG-Bench evaluation suite, focusing on tasks that pose challenges to existing language models. BBH consists of 23 tasks, and during the creation of the BBH dataset, researchers followed specific filtering criteria, including the number of task examples, task types, and performance of previous models. This dataset aims to advance the performance of language models on complex reasoning tasks and provides a valuable benchmark for future research efforts.

\subsection{Evaluation Metrics}
 Traditional approaches often employ string methods to determine if the output of an LLM is correct. Considering the method of extracting answers from LLM outputs using string methods and comparing them with correct answers may lead to the following issues:
\begin{itemize}
    \item High format requirements: This method requires precise formatting of LLM outputs, which may not always be consistent or predictable.
    \item Potential extraction of incorrect answers: LLMs may occasionally provide explanations for why incorrect answers are wrong, and extracting answers using string methods could inadvertently capture these explanations instead of the correct answers.
    \item Lack of definite correct answers: Many questions in natural language processing tasks do not have a single correct answer, making it challenging to determine the correctness of LLM outputs solely based on string matching.
\end{itemize}
Given these potential issues, relying solely on string extraction methods for answer evaluation may not be ideal, and alternative approaches, such as leveraging the judgment capabilities of the LLM itself, may be more suitable for accurate answer assessment.
\subsection{Results}
\paragraph{Ruozhiba.}

\begin{table}[h]
\centering
\begin{tabular}{ccc}
\toprule
Status & Count & Ratio \\
\midrule
Fail & 91 & 37.92\% \\
\midrule
Sucess & 149 & 62.08\% \\
\bottomrule
\end{tabular}
\caption{\textbf{Result of Ruozhiba}}
\label{tab:result of ruozhiba}
\end{table}

As shown in \cref{tab:result of ruozhiba}, Our framework achieved an accuracy of \textbf{62.08\%} on the Ruozhiba dataset. In fact, the Ruozhiba dataset poses a significant challenge to any natural language processing system due to its unique linguistic phenomena. The dataset is replete with Chinese-specific puns, ambiguities, and homophones, which are very common in the Chinese context but constitute a notable barrier for models like GPT-3.5-turbo, which are primarily trained on English corpora. Despite this, GPT-3.5-turbo has demonstrated commendable performance when dealing with the Ruozhiba dataset. This achievement not only proves the framework taps into LLM's powerful language understanding and generation capabilities but also shows its adaptability when faced with complex language structures. However, this accomplishment does not mean that GPT-3.5-turbo has fully mastered all the nuances of the Chinese language, and there are still limitations in its understanding and generation of Chinese content. Future research can continue exploring how to enhance the model's sensitivity and accuracy towards the Chinese context, as well as how to better utilize Chinese datasets for training and optimizing the model.

\paragraph{BIG-Bench-Hard.}

\begin{figure*}[ht]
    \centering
    \includegraphics[width=0.8\linewidth]{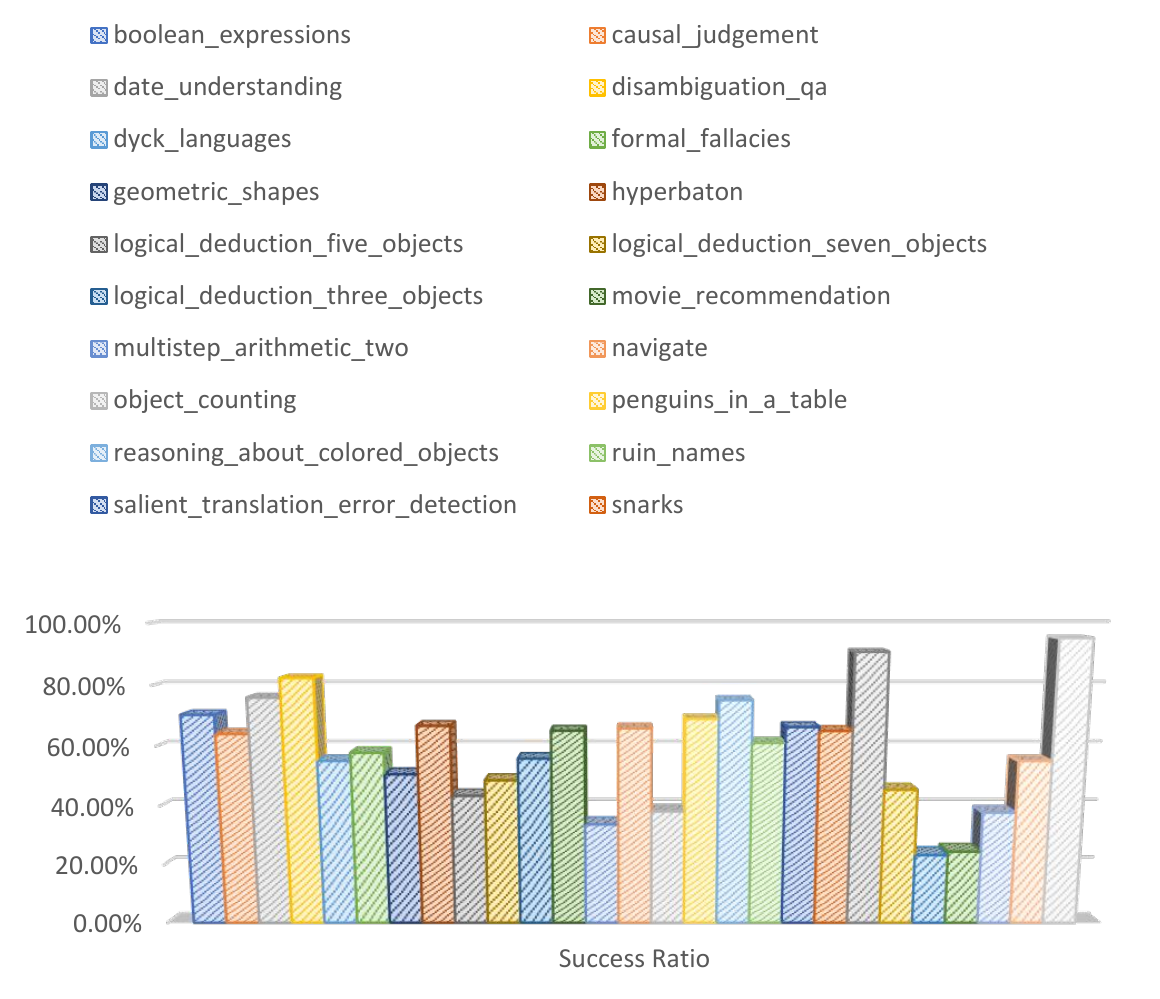}
	\caption{Result of BIG-Bench-Hard. This result includes 23 sub-tasks in BBH, a total of 27 sub-datasets, covering multiple aspects, with the job of determining whether the output answers are correct being accomplished by LLM.}
        \label{fig:BBH}
\end{figure*}

The BBH dataset incorporates knowledge from various domains such as world knowledge, natural language understanding, logical reasoning, and mathematics. As illustrated in \cref{fig:BBH}, the training results of our framework on the BBH dataset demonstrate outstanding performance in tasks related to world knowledge, natural language understanding, and logical reasoning. However, there is still room for improvement in handling mathematical problems and overly complex world knowledge.

\subsection{Ablation Study}
Our framework is composed of two integral components: an immutable component that guides the contemplation of the Large Language Models (LLMs) and a mutable component that is generated by the LLMs themselves. The primary focus of our experimental investigation is on the immutable component to substantiate the efficacy of the framework. This immutable component encompasses both Stimulating Prompt and Framework Prompt. Given that the construction of Framework prompts also integrates the principles of Stimulating Prompts, these Framework Prompts are indispensable and cannot be omitted.

Consequently, we conducted an experiment on the BBH dataset under identical settings, but with a crucial modification: we removed the Stimulating Prompts from the framework. These prompts, characterized by their uppercase formatting, actively encourage and steer the LLMs' thought processes. By eliminating these elements, we aimed to isolate and assess the impact of the Framework Prompts on the overall performance of the LLMs.

\begin{figure}[h]
    \includegraphics[width=\linewidth]{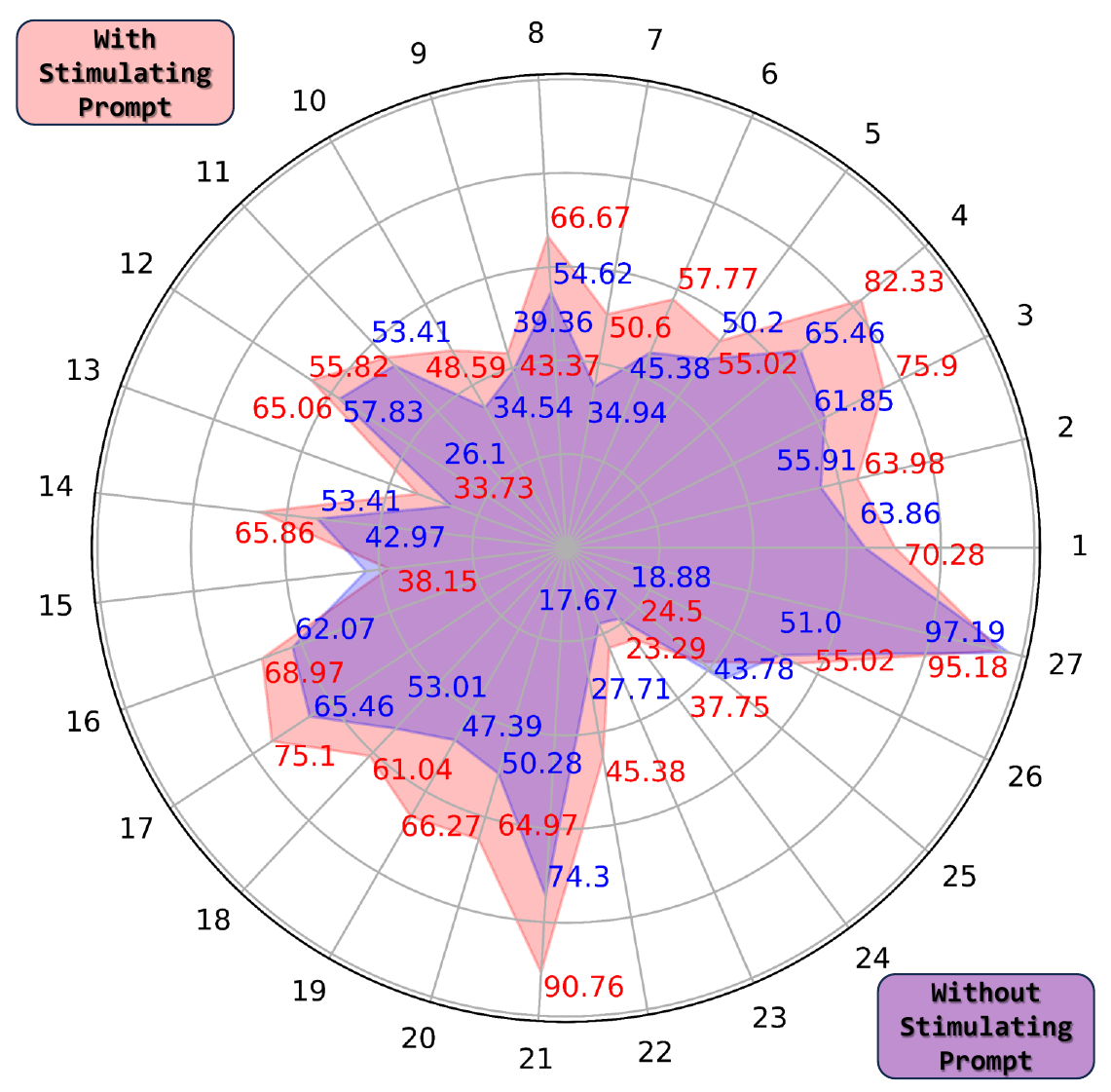}
	\caption{Comparison of results using Stimulating Prompts and without Stimulating Prompts on the BBH dataset.}
        \label{fig:radar_chart}
\end{figure}
From \cref{fig:radar_chart}, we can infer that, under identical conditions, the comprehensive effectiveness of using our framework surpasses that of not using it. This is sufficient evidence to validate the effectiveness of our framework.

\section{Conclusion}
\label{sec:Conclusion}
This study categorizes traditional prompts into two types: Stimulating Prompts and Framework Prompts. It then introduces a novel prompt that combines the advantages of both, automating its design with LLMs to form the Auto-Prompt Graphical Paradigm(APAG). An general Auto-Prompt Graphical Framework(APAF) is proposed as an instance of this paradigm, significantly enhancing the performance of Large Language Models (LLMs) in handling multi-domain issues. The framework fully leverages the strengths of both types of prompts, automating the prompt design process, guiding LLMs in conducting in-depth problem analysis, and optimizing solutions to ensure accuracy. Test results on the Ruozhiba and BBH datasets validate the framework's effectiveness, demonstrating LLMs' immense potential in complex problem-solving. Additionally, ablation studies confirm the efficacy of this paradigm. This success not only encapsulates the current state of prompt development but also introduces a new paradigm, illustrated with an example framework. Future work can further refine this paradigm, propose better frameworks, and greatly advance the application of LLMs.
\section{Limitations}
Based on the classification of prompts in this paper, we integrate the advantages of two types of prompts and achieve auto-prompt graphical paradigm design. Consequently, we propose a graph-based problem-solving framework that maximizes the positive response of LLMs to emotional stimuli. Additionally, we introduce a method capable of determining the correctness of LLM outputs on any dataset. Through experiments and ablation studies, we demonstrate the superior performance and effectiveness of the framework. However, our framework still has some shortcomings:
\begin{enumerate}
    \item The method of using LLM to judge the correctness of answers relies on the performance of the LLM, which may lead to misjudgments. However, proposing a task-specific evaluation method for each task does not align with our original intention of introducing a universal framework. In this context, we can opt to delegate the specific evaluation criteria to the LLM as well. This endeavor could further enhance the completeness of our framework, and we leave it to future work.
    \item We propose the current paradigm by drawing from existing Framework Prompts and Stimulating Prompts, along with our empirical insights. Through extensive experiments comparing with various potential frameworks, we have derived a relatively universal framework as an example. However, our experiments cannot cover every possible graph structure, which is practically impossible given the infinite nature of graph structures. Therefore, there are even more superior graph-based frameworks waiting for us to discover.
    \item In order to cover every scenario that LLM needs to handle, we have designed the framework to be as comprehensive as possible. Even when dealing with simple problems, the entire graph needs to be traversed thoroughly. While this approach ensures a thorough and exhaustive analysis of complex problems, it inevitably increases the cost of problem-solving for simpler tasks. To address this issue, we can propose a metric to evaluate the complexity of a problem, thereby determining whether to use our framework. This metric can be provided by the LLM. We can design a framework that contains both simple and complex sub-frameworks. Depending on the problem, the LLM can decide whether to use the complex framework or the simple one based on its judgment of the problem's complexity.
\end{enumerate}

Overall, there is still much room for optimization in our framework. This paradigm pioneers the automatic design of prompts that combine the advantages of two types of prompts in a graphical structure, offering a novel approach and providing a starting point for future work.

\clearpage


\bibliography{custom}




\end{document}